\documentclass[letterpaper, 10 pt, conference]{ieeeconf}
\usepackage{xcolor}
\usepackage{tablefootnote}

\usepackage{cite}
\usepackage{amsmath,amssymb,amsfonts}
\usepackage{algorithmic}
\usepackage{bm}
\usepackage{graphicx}
\usepackage{multirow}
\usepackage[utf8]{inputenc}
\usepackage[english]{babel}
\usepackage{tabularx,booktabs}
\newcolumntype{C}{>{\centering\arraybackslash}X} 
\usepackage{lipsum}
\usepackage{authblk}
\usepackage{textcomp}
\usepackage{amsmath}
\usepackage{xcolor}
\usepackage{subcaption}
\usepackage{booktabs}
\usepackage{array}
\usepackage[export]{adjustbox}
\title{\LARGE \bf Non-local Graph Convolutional Network for joint Activity Recognition and Motion Prediction}
\author{Dianhao Zhang$^1$, Ngo Anh Vien$^2$, Mien Van$^1$, Seán McLoone$^1$}
\affil{$^1$Queen's University Belfast, UK; $^2$Bosch Center for Artificial Intelligence, Germany}

\begin{document}

\maketitle
\thispagestyle{empty}
\pagestyle{empty}

\begin{abstract}
3D skeleton-based motion prediction and activity recognition are two interwoven tasks in human behaviour analysis. In this work, we propose a motion context modeling methodology that provides a new way to combine the advantages of both graph convolutional neural networks and recurrent neural networks for joint human motion prediction and activity recognition. 
Our approach is based on using an LSTM encoder-decoder and a non-local feature extraction attention mechanism to model the spatial correlation of human skeleton data and temporal correlation among motion frames. The proposed network can easily include two output branches, one for Activity Recognition and one for Future Motion Prediction, which can be jointly trained for enhanced performance.
Experimental results on Human 3.6M, CMU Mocap and NTU RGB-D datasets show that our proposed approach provides the best prediction capability among baseline LSTM-based methods, while achieving comparable performance to other state-of-the-art methods.   
\end{abstract}

\section{Introduction}
Human action recognition and motion prediction based on observations of their recent history of movement patterns and actions are vital capabilities that a robot must possess  in order to achieve safe and seamless human-robot collaboration (HRC). For both 3D skeleton-based action recognition and motion prediction, the key is to have a representation of human skeleton data that enables the extraction of informative features. Action recognition enables a robot to decide on the corresponding actions, while the anticipation of a human's future motion and position enables it to plan an execute its motions guaranteeing the safety of its human co-worker.


Developing a robotic system that can understand the circumstances under which it operates and reacts accordingly in a cooperative fashion is a challenging task. Especially for continuous HRC tasks containing multiple subtasks, it is necessary to enhance the context-aware skills of a robotic system by effectively modeling the historical human skeleton sequence. Recurrent and convolutional neural networks can extract features based on neighbour information in both the spatial and temporal domains. However, networks that focus on local interactions are often unable to adequately capture longer term dependencies. In this work, we address this {\bf first} challenge by capturing non-local spatial-temporal contexts via an application of a Graph Convolutional Network (GCN) based attention mechanism to enhance the performance of Long Short-Term Memory (LSTM) based sequence-to-sequence (Seq2Seq) models \cite{martinez2017human} for human motion prediction and activity recognition.  

The {\bf second} challenge for HRC systems is its computational complexity which needs to be minimized to enable practical (low-cost) real-time implementation. As human action recognition and motion prediction are separate tasks, the computation time and memory requirements to execute these two tasks simultaneously can be high. One way to achieve a more efficient implementation is to jointly train these two tasks in a single module so that they can consolidate each other's prediction. 
We propose a non-local GCN-based (NGC) mechanism to provide a high-level shared feature map to be shared between these two task network branches. The feature extraction module is based on the previously observed human behaviour including both short-term and long-term information. Our GCN based non-local feature extractor captures long-range dependencies directly by computing interactions between the hidden states of the encoder, regardless of the spatial distance between frames. The temporally long-range features help improve the motion prediction module, and provide informative feedback to enhance the activity recognition module. In particular, our two decoder branches are: 1) the human motion trajectory prediction module which utilizes a LSTM layer with residual connections to predict future human poses, and 2) the action recognition module which uses two feature extractors supported with feature augmentation and conditional random fields (CRFs) for activity classification. 


The main contributions of the paper are as follows:
\begin{itemize}
\item We propose a recurrent convolutional approach to capturing long-term dependencies enabling the motion prediction horizon based on non-local features to be increased. 
\item We apply a GCN based relational LSTM to provide shared spatial-temporal feature extraction for both human action recognition and motion prediction.   
\end{itemize}

\section{Related Work} 
\label{related}
\subsection{Neural Networks on Graphs}
Graph neural networks (GNN) \cite{scarselli2008graph} have received much attention recently as they can process data in irregular domains such as graphs or sets. In particular, graph convolution networks (GCN) \cite{kipf_semi-supervised_2017}, which are based on a fundamental convolution operation on the spectral domain, have been proven to be effective for processing structured data \cite{yan_spatial_2018-1}, and have been widely used to learn on structured skeleton data. 
Apart from graph recurrent neural networks (GRN) and graph convolutional neural networks (GCN), many alternative GNNs have been developed in the past few years, including graph autoencoders (GAEs) \cite{kipf2016variational} and spatial-temporal graph neural networks (STGNNs) \cite{yan_spatial_2018-1}. These learning frameworks can be built on GRN and GCN, or other neural architectures for graph modeling \cite{yan_spatial_2018-1}.

\subsection{Skeleton-based Human Action Recognition}
Many previous works have applied CNN or recurrent neural networks (RNN) \cite{liu_convolutional_2016, gao_optimized_2019} for action recognition on skeleton-based data. However, these approaches are limited in their capacity to learn the complex, irregular and non-Euclidean structure of skeleton data. One solution to this problem is to divide the human body into components and to extract the features for each component separately with subnetworks \cite{du2015hierarchical}. To further exploit the discriminative powers of different joints and frames, spatial-temporal attention \cite{liu2017global} can be employed in networks to enable them to selectively focus on discriminative joints of the skeleton within one frame, and pay different levels of attention to the outputs at different time instances.  

Recently, GNN or GCN have been demonstrated to be a more principled and effective choice for parsing the graph structure of skeleton data \cite{li_actional-structural_2019}, as they enable inner-correlation capture without segmentation of the whole body. Another example, TA-GCN \cite{heidari2020temporal}, uses an attention module in a GCN-based spatial-temporal model to extract more useful predictive features from graph data. In addition, to prevent the loss of information on the correlation between human body joints during training, \cite{peng2020learning} used a strategy called neural architecture search (NAS) at each iteration to enhance GCN-based human action recognition. 
\subsection{Skeleton-based Human Motion Prediction}
 Robust and fast human motion prediction in HRC can minimize robot response time and latency, thereby improving task performance.
%
%
With the advances in deep learning, RNN based approaches such as LSTM have been shown to be a powerful tool for motion prediction in recent years \cite{liu_deep_2019}. A recent common strategy for motion prediction is to use a recurrent neural network (RNN) to encode temporal information \cite{martinez2017human, fragkiadaki2015recurrent}. Although sequence-to-sequence models perform well for short-term prediction, encoding of long-term historical information is challenging \cite{bahdanau2014neural}. To account for inter-dependency between connected human body components, graph based approaches are applied to learn the structural dynamics of the different component interactions. Spatial-temporal graph neural networks (STGNNs) have been proposed to model such systems \cite{wu_comprehensive_2020} considering inner spatial-temporal information.  
\section{Spatial-Temporal Graph Neural Network Based Human Feature Learning}
As a summary, our network architecture will consist of the following modules:


(1) {\bf A GCN-LSTM Encoder} which models human motion dynamics by encoding the temporal evolution of skeletal graph data in an encoding vector. We employ GCN layers to extract features from raw skeletal graphs, followed by bi-directional LSTM (Bi-LSTM) layers to learn past and future contexts within a sequence of $\tau$ poses.

(2) {\bf A Non-local GCN Module} which bridges the encoded state of the GCN-LSTM Encoder with the action recognition and motion prediction modules. We employ residual blocks (Res-GCN) \cite{mao_learning_2019} to extract spatial features from each frame, while the GCN-based non-local network models the relationships between the current hidden state at time $t$ and the hidden states from $t=1$ to $t=\tau$ using a self-attention architecture. 

(3) {\bf Two Task-specific Decoders}: The first LSTM-based decoder is designed to predict the future sequence of skeletal graph poses (human motion prediction). The second CNN-CRF-based decoder performs action recognition.


\subsection{GCN-LSTM Encoder}
\begin{figure}[thpb]
\centering
\noindent\includegraphics[width=\columnwidth]{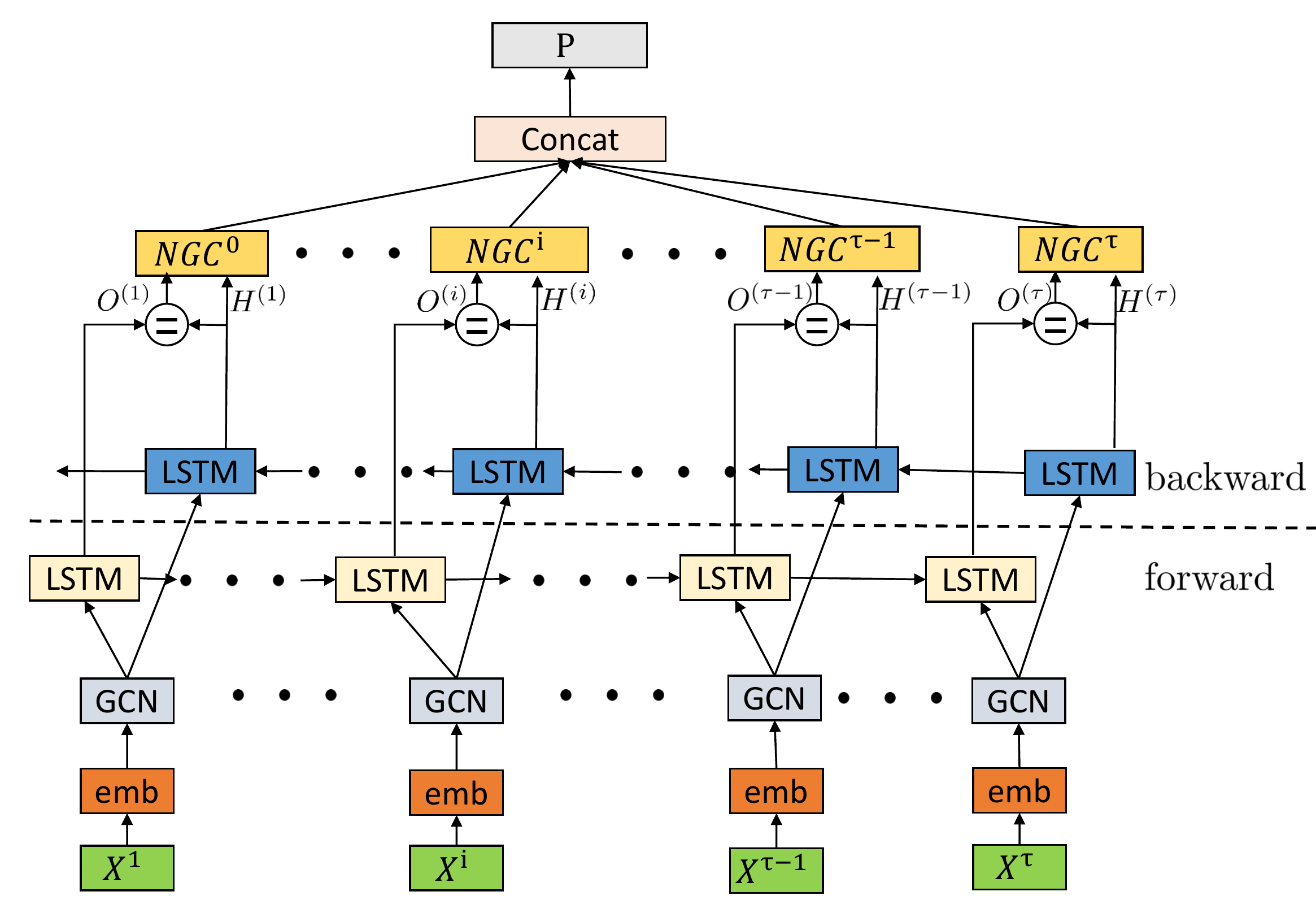}
\caption{Architecture of the GCN-LSTM encoder module and how it interfaces with the NGC module. The input to an NGC block includes the output and hidden states $\{O,H\}$ generated from the Bi-LSTM layers.}
\label{fig1}
\end{figure}

The overall structure and information flow of our encoder module are shown in Fig. \ref{fig1}. In the first layer of the encoder, the skeleton graph is embedded in a vector $E \in \mathbb{R}^{N \times 3}$ by a fully connected (FC) layer, where $N$ denotes the number of human joints. The input to the embedding layer is $X_{prev}$,  where $X_{prev} = [\textbf{X}^{(1)}, \cdots, \textbf{X}^{(\tau)}] \in \mathbb{R}^{\tau \times N \times 3}$ is a sequence of $\tau $ skeletal graphs. This is then fed into a GCN layer to extract spatial graph feature $G$. Then we apply a 4-layer Bi-LSTM to generate output $O$ and hidden states $H$. Denoting $f_{\rm{em}}, f_{\rm{GCN}}, f_{\rm{Bi-LSTM}}$ as the embedding, GCN and Bi-LSTM encoder functions, respectively, the operation of the encoder module can be formulated mathematically as:
\begin{align}
    E &= f_{\rm{em}} (X_{prev}; \boldsymbol{\theta}_{emb})\\
    G &= f_{\rm{GCN}}(E; \boldsymbol{\theta}_{GCN})\\
    O, H &= f_{\rm{Bi-LSTM}} (G; \boldsymbol{\theta}_{\rm{Bi-LSTM}})
\end{align}
where $E = \{\epsilon_1, \cdots, \epsilon_{\tau}\}$ is the matrix of embedded pose feature output from the fully connected layer for the observed sequence of $\tau$ skeleton frames. Parameters $\boldsymbol{\theta}_{emb}$, $\boldsymbol{\theta}_{GCN}$ and $\boldsymbol{\theta}_{\rm{Bi-LSTM}}$ are the trainable parameters of the fully connected embedding layer, GCN layer and Bi-LSTM layer, respectively. 

\subsection{Non-local GCN Module}
\begin{figure}[thpb]
\centering
\noindent\includegraphics[width=\columnwidth]{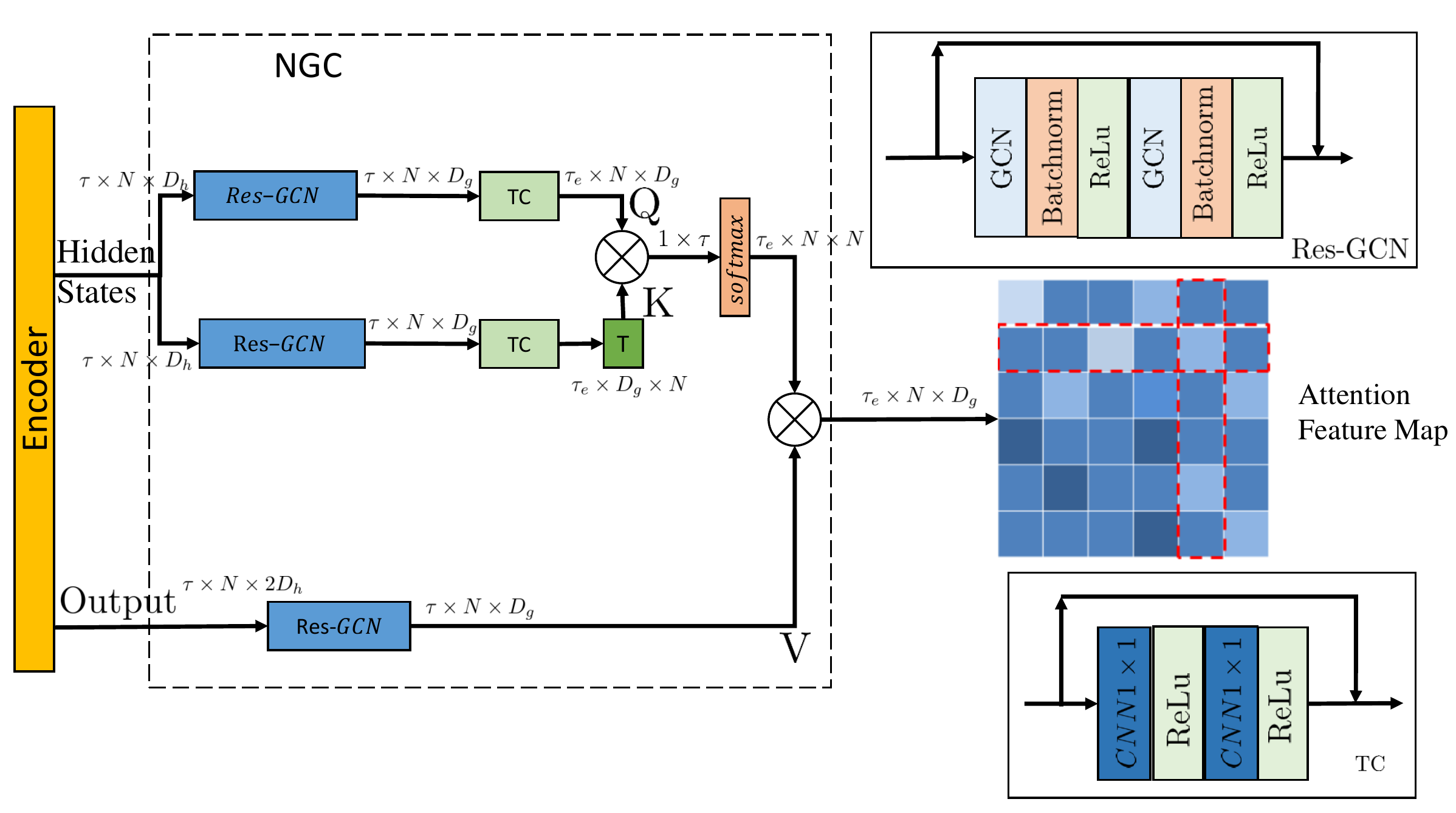}
\caption{A non-local GCN (NGC) based feature extractor: The GCN based self-attention module generates an attention containing spatial-temporal importance according to the hidden states of a recurrent network. The upper two channels in the self-attention module compute the relationship between hidden states by using Res-GCNs and temporal convolution (TC) operations. The generated matrix representing the importance of frames is multiplied by the extracted output feature from the encoder to get an attention matrix.}
\label{fig2}
\end{figure}

In this work, we employ two GCN layers that are based on residual blocks (Res-GCN) \cite{mao_learning_2019} and the self-attention mechanism \cite{vaswani_attention_2017} to further extract non-local features from the output of the encoder, as described in Fig. \ref{fig2}. The attention block processes three inputs, a query $\emph{Q}$, keys $\emph{K}$, and values $\emph{V}$  generated from the hidden states and output of the encoder block. The output of the self-attention function is defined as:
\begin{align}
    Att(\emph{Q}, \emph{K}, \emph{V}) = \frac{1}{Z}\mathrm{softmax}({\emph{Q} \emph{K}^T})\emph{V},
\end{align}
where $Z$ is a normalization factor.
The query and the key are extracted from the hidden states within observed time length $\tau$, $\emph{Q}, \emph{K} \in \mathbb{R}^{\tau \times N \times D_h}$ and $N_h = N \times D_h$ is the hidden size of the Bi-LSTM layer from the encoder. The value $\emph{V}$ is achieved by applying Res-GCN function on encoder output, $\emph{V} \in \mathbb{R}^{\tau \times N \times 2D_h}$, where $N_{out} = N \times 2D_h$ denotes the output size of the encoder. The spatial-temporal correlation between $\emph{Q}$ and $\emph{K}$ is calculated and added into the generated feature map after multiplying with $\emph{V}$. 

The input skeleton to the GCN is modeled as a fully-connected graph with $N$ nodes. Firstly, a Res-GCN with trainable weights $W^{(t)} \in R^{N_h \times N_g}$ ($N_g = N \times D_g$ is the output dimension of Res-GCN layer) processes as inputs the hidden state matrix $H \in R^{\tau \times N \times D_h}$. Then a temporal convolution operation is employed for temporal feature extraction after the Res-GCN layer. The process by which ResGCN-TC generates $\emph{Q}$ can be described as:
\begin{align}
   g(H) = \mathrm{ReLU}(f_{bn}(\bm{\hat{D}}^{-\frac{1}{2}} \bm{\hat{A}} \hat{\bm{D}}^{-\frac{1}{2}} H W)),\\
   f_{RG} (\cdot) = \mathrm{ReLU}(f_{bn}(\bm{\hat{D}}^{-\frac{1}{2}} \bm{\hat{A}} \hat{\bm{D}}^{-\frac{1}{2}} g(\cdot) W)),\\
   f_{TC}(\cdot) = \mathrm{ReLU}(\mathrm{Conv}(\mathrm{ReLU}(\mathrm{Conv}(\cdot)))),
\end{align}
where $f_{bn}(\cdot)$ is the batch normalization function and $\emph{Q} = f_{TC}(f_{RG}(H))$ is the Res-GCN function to achieve query $\emph{Q}$. The functions $f_{RG}$ and $f_{TC}$ denote the Res-GCN function and temporal convolution function, respectively. The Res-GCN module consists of two GCN layers, each followed by batch normalization and ReLU activation, while the TC module consists of two CNN layers, each followed by a ReLU activation function, as described in Fig. \ref{fig2}. The term $D$ is the diagonal node degree matrix used to normalize the adjacency matrix A, and $\mathrm{ReLU}(\cdot)$ is the $\mathrm{ReLU}$ activation function. In practice, a symmetric normalization $\textbf{D}^{-\frac{1}{2}} \textbf{A} \bm{D}^{-\frac{1}{2}}$ is used to avoid merely averaging of neighboring nodes. In this work, $\bm{\hat{A}} = \bm{A} + \bm{I}$ so that a self-loop is included. $\bm{I}$ is the identity matrix and $\bm{\hat{D}}$ is the diagonal node degree matrix of $\bm{\hat{A}}$. After the Res-GCN operation, $\emph{Q}$ and $\emph{K}$ are transformed to dimension $\tau \times N \times D_g $. A TC layer applies a convolutional operation over the temporal dimension to generate a spatial-temporal feature map. The generated spatial-temporal feature matrices are multiplied together and passed through a softmax function to capture spatial-temporal correlation between hidden states.  We can describe this non-local feature extraction process as:
\begin{align}
f &= f_{TC}(f_{RG}(.))\\
P &= \mathrm{softmax}(f(H) f(H)^T) f_{RG}(O) 
\label{featuremap}
\end{align}

\subsection{Activity Recognition in Human Robot Collaboration}
\begin{figure}[thpb]
\centering
\noindent\includegraphics[width=\columnwidth]{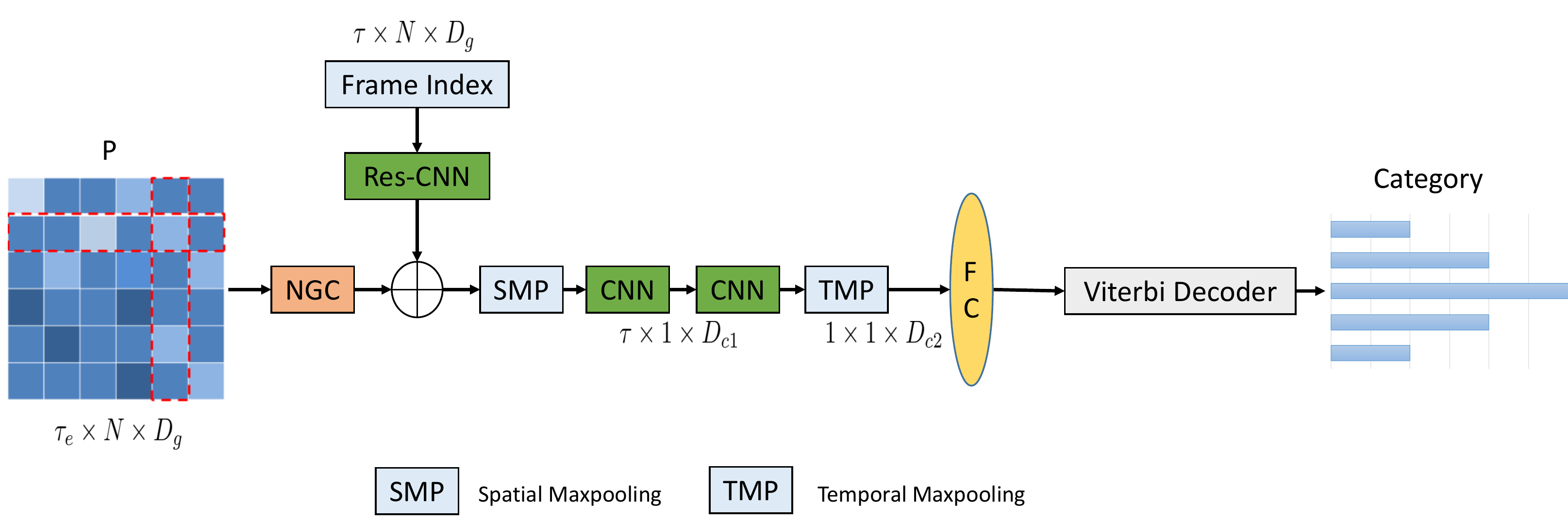}
\caption{NGC based model for action recognition. Before importing the feature map to each NGC block, a feature augmentation operation is introduced. Then, conditional random fields are applied to the generated feature map to get the action categories.}
\label{fig6}
\end{figure}

Given the sophisticated attention feature map $P$ in Eq. \ref{featuremap} extracted from observed human skeleton data, we now propose the first decoder to classify activity. We first augment this feature map by incorporating the temporal semantics extracted by a Res-CNN layer, that is, a semantic feature is extracted from the spatial-temporal frame index sequence using an Res-CNN and added to the extracted attention feature map, $P$. Then we apply a spatial temporal pooling (SMP) operation on the generated feature map to obtain a temporal feature representation. Then two temporal convolutional neural network layers combined with max-pooling are employed to learn the feature representation for classification. In this work, we employ CRFs for label sequence estimation \cite{fraccaro_disentangled_nodate}. CRFs model the conditional probability of a label sequence $Y$, given an extracted attention feature of observed frames, $P^*$, as:
\begin{align}
P(Y|P^*) = \frac{1}{Z(P^*)} \prod_{t=1}^{\tau} \Psi_f (y_{t-1}, y_t, P_t^*)
\end{align}
where $\tau$ denotes the observed time length, and $\Psi_t$ is the weight function on the transition from state $y_{t-1}$ to state $y_t$ when the current observation is $P_t^*$. Feature map $P^*$ contains the full body features in observed frames $\{1,\ldots,\tau\}$, while $Z$ is a normalization function. The goal of the proposed model, as depicted in Fig. \ref{fig6}, is for a given feature map $P^*$ and the model's parameter vector $\theta$, to find the most probable label $y_t$ by maximizing the conditional probability $P(y_t|P^*; \theta)$. This procedure is implemented by a Viterbi decoder to  estimate the probability of the most probable path ending.

\subsection{NGC Attention based LSTM Encoder-Decoder for Human Motion Prediction}
The output of the NGC feature extractor represents a context that can be used as the initial hidden state for the motion prediction decoder. The first input sequence to the decoder is the first frame in $X^{(\tau + 1)}$, $f^{(0)}$.  In the subsequent steps of decoding, $f^{(t-1)}$  is the decoder’s own prediction of the previous step or the ground-truth of the previous step. The teacher forcing method is applied to address slow convergence and instability when training recurrent networks.  Based on this input representation, the decoder generates an output sequence that represents a sequence of future motion. 

\begin{figure}[thpb]
\centering
\noindent\includegraphics[width=\columnwidth]{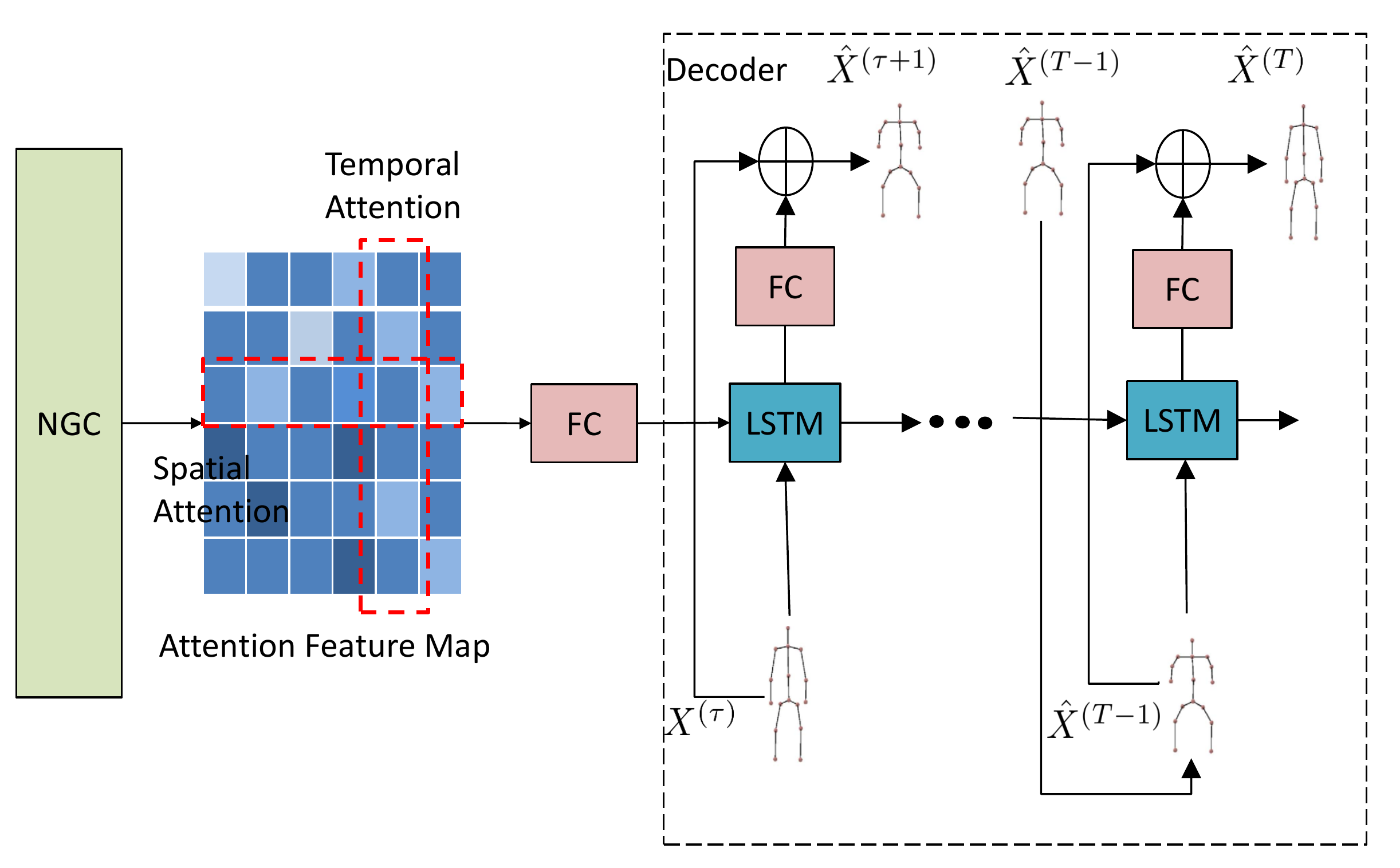}
\caption{LSTM decoder for human motion prediction}
\label{fig5}
\end{figure}

Our LSTM based motion prediction module, depicted in Fig. \ref{fig5}, is inspired by the Seq2Seq network \cite{sutskever2014sequence}, so we name it NGC-Seq2Seq. It computes future motion predictions as $\hat{X}_{fut}= f_{pred}(X_{prev};\theta_{pred})$, where we denote $f_{pred}(.)$ as the motion prediction module, $X_{fut}= [X^{\tau + 1}, ... , X^{T}]$ as the future motion sequence; and $\theta_{pred}$ as the trainable parameters of the prediction module. In particular, to produce the $(t+1)^{th}$ pose $(t \geq 0)$ the motion prediction module performs the following operations:
\begin{align}
H_{F}^{(\tau + 1)} &= f_{LSTM}(\Delta {P}, \bm{\theta}_{pred}), \\
\hat{X}_{fut}^{(\tau + 1)} &= f_{FC}(\Tilde{H}_{F}^{(\tau+1)}) + P,\\
\hat{X}_{fut}^{(\tau+2)} &=\hat{X}_{fut}^{(\tau+1)} + f_{LSTM}(\Delta \hat{X}_{fut}^{\tau + 1}, \bm{\theta}_{pred})
\end{align}
Here, $f_{LSTM}(.)$ and $f_{FC}(.)$ represent LSTM decoder predictor and fully connected layers (FC), respectively. As we are predicting an offset between frames, the term $\Delta X_F^{(t)}$ denotes the displacement of the current human motion sequence at time $t$, $\Tilde{H}_{fut}^{(t)}$ is the hidden state of the LSTM at time $t$, and $\Delta {P}$ denotes the displacement of the attention feature map, $P$, between the current $\tau$ and the previous $(\tau-1)$ prediction steps. Firstly, we feed the updated hidden states and current feature displacement into the LSTM decoder to produce the features that reflect future displacement. The first input to the LSTM decoder is the last output from the attention feature $P$. Then, we adopt the predicted displacement over the previous pose to predict the next frame. 
One limitation for the LSTM decoder is, during inference, the previous true target tokens are replaced by tokens generated from the model's hidden state \cite{bengio2015scheduled}. To solve this problem for better long-term anticipation performance, we apply teacher forcing as a learning strategy to diminish the influence from the model's own error \cite{ranzato2015sequence}. At every time step, the choice on ground truth or its own prediction is determined by a coin flipping probability $p$. Initially, $p = 1$ (i.e. teacher forcing), and it decays exponentially with a factor $\beta = 0.995$ per epoch. 

\section{EXPERIMENTS AND ANALYSIS }

\subsection{Implementation Details and Metrics}
The models are implemented with PyTorch. 
The size of the input to the encoder is $N \times 3$, and the hidden size of the four-layer Bi-LSTM is $N \times 8$. In the NGC module, for the channels receiving hidden states,  the first GCN layer in the Res-GCN converts the spatial dimension from $N \times 8$ to $N \times 4$, while the second GCN layer converts the spatial dimension from $N \times 4$ to $N \times 3$. As the output from the encoder is the concatenation of the forward and backward hidden states, in terms of the Res-GCN on the bottom channel, the first GCN layer converts the spatial dimension from $N \times 16$ to $N \times 8$, while the second GCN layer converts from $N \times 8$ to $N \times 3$. The TC in the bottom channel is the same as the ones in previous channels. The kernel size of the temporal convolution is $1 \times 1$. The first layer of the CNN in the TC converts the temporal dimension from $\tau$ to 64, while the second CNN layer converts the temporal dimension from 64 to $\tau$.

In the motion prediction decoder, the hidden size of the LSTM decoder is $N \times 8$ and the fully connected layer converts the spatial dimension back to $N \times 3$, which can be formed as skeleton data. In the action recognition decoder, the Res-CNN used to process temporal semantics converts the temporal dimension from $\tau$ to 64, then back to $\tau$. The two-layer CNN between spatial-temporal max-pooling converts the spatial dimension from $N \times 8$ to $N \times 16$. The loss used for joint angles is the $l_1$ average distance between the ground-truth joint angles, and the predicted ones. We apply the  Huber loss \cite{huber1992robust} as the prediction loss. We also include a penalty term in the Huber loss, $p = \gamma_p(x^2+y^2+z^2-1)^2$, where $x$, $y$ and $z$ are the 3D coordinates of the predicted and ground truth skeleton, and $\gamma_p=0.1$ is the penalty weighting. Denoting the $i^{th}$ predictions and ground truth as $(X_{fut})_i$ and $(\hat{X}_{fut})_i$, for $N_T$ samples in one mini-batch, the prediction loss $L_{pred}$ can be expressed as:
\begin{align}
L_{pred}^i = \frac{1}{N_T} 
   \begin{cases}
   \frac{0.5(X_{fut}^i - \hat{X}_{fut}^i)^2}{\beta} & \text{if $X_{fut}^i - \hat{X}_{fut}^i < \beta$} \\
   |X_{fut}^i - \hat{X}_{fut}^i|- 0.5\beta & \text{otherwise}
   \end{cases}
\end{align}
For action recognition, the conditional random field are commonly trained by maximizing the conditional log likelihood of a labeled training set to estimate the weight vector. Let the true label of the $n^{th}$ sample $(X_{fut})_n$ be $Y_n$ and the estimated be $\hat{Y}_n$. For $N_T$ training samples in one mini-batch, the action recognition loss is formulated as:
\begin{align}
\label{eqn:eqlabel}
\begin{split}
L_{rec} = \min_{\bm{\omega}} \left\{ \frac{\alpha}{2} \Vert \omega^2 \Vert + \frac{1}{N_T} \sum_{n=1}^{N_T} L_i (Y, \hat{Y}) \right\},
\end{split}
\end{align}
where the factor $\alpha$ is a trade-off constant, which can be exploited to provide a regularization balance term between the model complexity and fitting the data. It is used to avoid or reduce over-fitting in model learning. Parameter $\bm{\omega}$ is the trainable parameter of the CRF model. Inspired by \cite{liu2020double}, the function $L_n (Y, \hat{Y})$ denotes the measure cost of the wrong estimation for the $n^{th}$ training sample, which is expressed as:
\begin{align}
  L_n(Y, \hat{Y}) &= \frac{1}{T-\tau} \sum_{t=1}^{T-\tau} (1- \delta (Y_t^n, \hat{Y}_t^n))\\
  \text{where }\delta (Y_t^n, \hat{Y}_t^n) &= 
  \begin{cases}
   1 & \text{$Y_t^n = \hat{Y}_t^n$}\\
   0 & \text{otherwise}
  \end{cases}
\end{align}

We can jointly train the motion prediction and activity recognition modules with a combined loss function as:
\begin{align}
    L = \lambda L_{pred} + (1-\lambda) L_{rec},
    \label{combinedloss}
\end{align}
where $\lambda$ is a trade-off value to balance the importance of the two tasks. We have carried out a grid-search to optimise $\lambda$, and found that $\lambda = 0.4$ yields the best validation loss. We use the Adam optimizer to train our model with the learning rate initially set as 0.001 and then decayed by a factor of 10 every 10 epochs. The model is trained with batch size 16 for 100 epochs on a NVIDIA Quadro P4000 GPU. 
\subsection{Datasets}
\paragraph{Human 3.6M dataset}
Human 3.6M is a dataset containing 3.6 Million accurate recordings of 3D Human poses of motion \cite{ionescu_human36m_2014}. The recordings include the skeleton data for 32 joints.

\paragraph{NTU RGB+D Dataset}
The NTU-RGB+D was introduced by Amir Shahroudy et al. \cite{shahroudy_ntu_2016} in 2016. The dataset consists of 56, 880 RGB+D video samples, captured from 40 different human subjects. The dataset included RGB videos, depth sequences, skeleton data (3D locations of 32 major body joints), and infrared frames. This dataset contains both human motion and labeled activity. NTU RGB+D also provides two standard test methods: Cross subject (CS) and Cross view (CV). 

\paragraph{CMU Mocap}
The CMU Mocap dataset was collected in a lab using 12 Vicon infrared MX-40 cameras and includes skeleton data for 41 joints. There are 144 human motion categories in this datatset. 

\subsection{3D Skeleton-based Action Recognition}
For action recognition, we first show the classification accuracy of our method in comparison with a number of baselines on two NTU-RGB+D benchmarks, i.e. Cross-Subject (CS) and Cross-View (CV). Table \ref{table1} presents the recognition accuracy of the methods considered, and shows that the proposed NGC-CRF model outperforms all baselines on both benchmarks. Note that our network is trained to exploit the motion prediction module, that is, it uses both human motion and labeled activity data.
\begin{table}[h]
\centering
\begin{tabular}{|l|l|l|p{0.5cm}|}
\hline
Methods       & CS     & CV     \\ \hline
Lie Group \cite{vemulapalli_human_2014}     & 50.1\% & 52.8\% \\ 
H-RNN \cite{yong_du_hierarchical_2015}         & 59.1\% & 64.0\% \\ 
Deep LSTM \cite{shahroudy_ntu_2016}     & 60.7\% & 67.3\% \\ 
PA-LSTM \cite{shahroudy_ntu_2016}       & 62.9\% & 70.3\% \\ 
ST-LSTM+TS \cite{liu2016spatio}    & 69.2\% & 77.7\% \\ 
Temporal Conv \cite{kim2017interpretable} & 74.3\% & 83.1\% \\ 
Visualize CNN \cite{dong_action_2020} & 76.0\% & 82.6\% \\ 
ST-GCN \cite{yan_spatial_2018-1}        & 81.5\% & 88.3\% \\ 
DPRL \cite{tang_deep_2018-1}          & 83.5\% & 89.8\% \\ 
SR-TSL \cite{si_skeleton-based_2018}        & 84.8\% & 92.4\% \\ 
HCN \cite{li_co-occurrence_2018}           & 86.5\% & 91.1\% \\ 
STGR-GCN \cite{li_spatio-temporal_2019}      & 86.9\% & 92.3\% \\ 
motif-GCN \cite{wen_graph_2019}     & 84.2\% & 90.2\% \\ 
AS-GCN \cite{li_actional-structural_2019}        & 86.8\% & 94.2\% \\ 
AGC-LSTM \cite{si_attention_2019-1}      & 88.5\% & 95.1\% \\ 
DGNN \cite{shi_skeleton-based_2019-1}          & 89.2\% & 95.0\% \\ 
Sym-GNN \cite{li_symbiotic_2019}         & 90.1\% & 96.4\% \\ 
\hline
NGC-CRF     & \textbf{95.2\%} & \textbf{98.5\%} \\ 
\hline
\end{tabular}
\caption{Comparison of action recognition on NTU-RGB+D. The accuracy on
both Cross-Subject (CS) and Cross-View (CV) benchmarks.}
\label{table1}
\end{table}

\subsection{3D Skeleton-based Motion Prediction}
To validate the proposed model, we show the prediction performance for both short-term and long-term motion prediction on Human 3.6M (H3.6M). We quantitatively evaluate various methods in terms of the mean angle error (MAE) between the generated motions and ground-truths in angle space. 
\paragraph{Short-term motion prediction} 
Short-term motion prediction aims to predict the future poses over 400 milliseconds. We train our model to generate the future 10 frames for each input sequence on Human 3.6M. The results in Table \ref{tab:addlabel1} show that our model can capture long-term dependencies to achieve stable short term prediction. While, the performance of the model over the shortest time scale (80 ms) is worse than Imit-L, Traj-GCN, and Sym-GNN, it is competitive with the other methods on the longer time scales and provides the best performance when predicting poses at 400 ms, which is the length of the previous human motion sequences used as input to the model.
\begin{table*}[h]
  \centering
  \caption{Comparisons of MAEs between NGC-Seq2Seq and state-of-the-art methods for short-term motion prediction on four representative actions of H3.6M.}
  \small\addtolength{\tabcolsep}{-5pt}
  \resizebox{\linewidth}{!}{%
    \begin{tabular}{|c|cccc|cccc|cccc|cccc|}
    \hline
    Motion & \multicolumn{4}{c|}{Walking}  & \multicolumn{4}{c|}{Eating}   & \multicolumn{4}{c|}{Smoking}  & \multicolumn{4}{c|}{Discussion} \\
    \hline
    Milliseconds  & 80    & 160   & 320   & 400   & 80    & 160   & 320   & 400   & 80    & 160   & 320   & 400   & 80    & 160   & 320   & 400 \\
    \hline
    ZeroV \cite{martinez2017human} & 0.39  & 0.68  & 0.99  & 1.15  & 0.27  & 0.48  & 0.73  & 0.86  & 0.26  & 0.48  & 0.97  & 0.95  & 0.31  & 0.67  & 0.94  & 1.04  \\
    Res-sup \cite{martinez2017human} & 0.27  & 0.46  & 0.67  & 0.75  & 2.23  & 0.37  & 0.59  & 0.73  & 0.32  & 0.59  & 1.01  & 1.10  & 0.30  & 0.67  & 0.98  & 1.06  \\
    CSM \cite{li_convolutional_2018} & 0.33  & 0.54  & 0.68  & 0.73  & 0.22  & 0.36  & 0.58  & 0.71  & 0.26  & 0.49  & 0.96  & 0.92  & 0.32  & 0.67  & 0.94  & 1.01  \\
    TP-RNN \cite{chiu2019action} & 0.25  & 0.41  & 0.58  & 0.65  & 0.20  & 0.33  & 0.53  & 0.67  & 0.26  & 0.47  & 0.88  & 0.90  & 0.30  & 0.66  & 0.96  & 1.04  \\
    AGED \cite{ferrari_adversarial_2018} & 0.21  & 0.35  & 0.55  & 0.64  & 0.18  & 0.28  & 0.50  & 0.63  & 0.27  & 0.43  & 0.81  & 0.83  & 0.26  & 0.56  & 0.77  & 0.84  \\
    Skel-TNet\cite{guo2019human} & 0.31  & 0.50  & 0.69  & 0.76  & 0.20  & 0.31  & 0.53  & 0.69  & 0.25  & 0.50  & 0.93  & 0.89  & 0.30  & 0.64  & 0.89  & 0.98  \\
    Imit-L \cite{aksan2020spatio} & 0.21  & 0.34  & 0.53  & 0.59  & 0.17  & 0.30  & 0.52  & 0.65  & 0.23  & 0.44  & 0.87  & 0.85  & 0.23  & 0.56  & 0.82  & 0.91  \\
    Traj-GCN \cite{mao_learning_2019} & 0.18  & 0.32  & 0.49  & 0.56  & 0.17  & 0.31  & 0.52  & 0.62  & 0.22  & 0.41  & 0.84  & 0.79  & $\bm{0.20}$  & 0.51  & 0.79  & 0.86  \\
    Sym-GNN \cite{li_symbiotic_2019} & $\bm{0.17}$  & 0.31  &  0.50  &  0.60  & $\bm{0.16}$  & 0.29  & 0.48  & 0.60  & $\bm{0.21}$  & 0.40  &  0.76  & 0.80  & $\bm{0.21}$  & 0.55  & 0.77  & 0.85  \\
    \hline    
    Without NGC & 0.33  & 0.54  & 0.78  & 0.91  & 0.28  & 0.45  & 0.65  & 0.83  & 0.35  & 0.62  & 1.03  & 1.14  & 0.35  & 0.71  & 1.01  & 1.09  \\
    \end{tabular}}%
  \label{tab:addlabel1}%
\end{table*}%
\paragraph{Long-term motion prediction} Here we define long-term motion prediction as predicting poses over a 1000 milliseconds horizon. In this work, we compare the long-term performance with state-of-the-art baselines on the CMU Mocap dataset in Table \ref{tab:addlabel2}. The experimental results show that the proposed model substantially exceeds the baselines in both short-term and long-term prediction. 
\begin{table*}[h]
  \centering
  \setlength{\tabcolsep}{0.3cm}
  \caption{Comparisons of MAEs between NGC-Seq2Seq and state-of-the-art methods for long-term motion prediction on three representative actions of CMU-Mocap.}
  \resizebox{\linewidth}{!}{%
    \begin{tabular}{|l|rrrrr|rrrrr|rrrrr|}
    \hline
    Motion & \multicolumn{5}{c|}{Basketball}       & \multicolumn{5}{c|}{Jumping}          & \multicolumn{5}{c|}{Direction Traffic} \\
    \hline
    Milliseconds & 80    & 160   & 320   & 400   & 1000  & 80    & 160   & 320   & 400   & 1000  & 80    & 160   & 320   & 400   & 1000 \\
    \hline
    Res-sup \cite{martinez2017human} & 0.49  & 0.77  & 1.26  & 1.45  & 1.77  & 0.32  & 0.67  & 0.98  & 1.06  & 2.40  & 0.31  & 0.58  & 0.94  & 1.10  & 2.06  \\
    Skel-Tnet \cite{guo2019human} & 0.35  & 0.63  & 1.04  & 1.14  & 1.78  & 0.30  & 0.64  & 0.89  & 0.98  & 1.99  & 0.22  & 0.44  & 0.78  & 0.90  & 1.88  \\
    Traj-GCN \cite{mao_learning_2019} & 0.33  & 0.52  & 0.89  & 1.06  & 1.71  & 0.20  & 0.51  & 0.79  & 0.86  & 1.80  & \textbf{0.15}  & 0.32  & 0.52  & 0.60  & 2.00  \\
    Sym-GNN \cite{li_symbiotic_2019} & 0.32  & 0.48  & 0.91  & 1.06  & 1.47  & 0.32  & 0.55  & 1.40  & 1.60  & 1.82  & 0.20  & 0.41  & 0.75  & 0.87  & 1.84  \\
    \hline
    Without NGC & 0.32  & 0.54  & 0.77  & 0.89  & 1.77  & 0.61  & 0.92  & 1.45  & 1.79  & 1.98  & 0.52  & 1.01  & 1.56  & 1.74  & 2.03  \\
    \textbf{NGC-Seq2Seq} & \textbf{0.27 } & \textbf{0.45 } & \textbf{0.67 } & \textbf{0.82 } & \textbf{1.41 } & \textbf{0.16 } & \textbf{0.45 } & \textbf{0.61 } & \textbf{0.82 } & \textbf{1.46 } & \textbf{0.15}  & \textbf{0.27 } & \textbf{0.49 } & \textbf{0.57 } & \textbf{1.53 } \\
    \hline
    \end{tabular}}%
  \label{tab:addlabel2}%
\end{table*}%

\section{Ablation Study}
\subsection{Mutual Effects of Prediction and Recognition}
As motion prediction and action recognition are trained simultaneously, a combined loss (Eq.18) is presented to train the HRC system. In order to achieve action category and predicted human pose simultaneously, we train the model on the NTU-RGB+D dataset for different values of $\lambda$, where $\lambda$ is the trade-off value applied between motion prediction and action recognition losses. Table IV presents the recognition accuracy and average MAEs for short-term prediction on NTU RGB+D and Human 3.6M, respectively. The results suggest that appropriate combinations of the prediction and recognition loss can generate a more stable HRC system. According to Table IV, choosing $\lambda = 0.4$ is most effective for both recognition and prediction. Compared to models trained with separate loss functions, the weighted combination achieves better performance and enables simultaneous training of both prediction modules. 

\begin{table}[h]
  \centering
  \caption{Action recognition accuracy on NTU-RGBD and average MAEs on Human 3.6M with different $\lambda$.}
  \small\addtolength{\tabcolsep}{-5pt}
  \resizebox{\linewidth}{!}{%
    \begin{tabular}{|c|c|c|c|c|c|c|}
    \hline
    \multirow{2}[2]{*}{$\lambda$} & \multirow{2}[2]{*}{CS} & \multirow{2}[2]{*}{CV} & \multicolumn{4}{c|}{MAEs} \\
          &       &       & \multicolumn{1}{c}{80 ms} & \multicolumn{1}{c}{160 ms} & \multicolumn{1}{c}{320 ms} & 400 ms \\
    \hline
    0     & 95.20\% & 98.50\% & 0.22  & 0.36  & 0.47  & 0.63 \\
    0.2   & 90.59\% & 94.11\% & 0.21  & 0.33  & 0.48  & 0.65 \\
    0.3   & 79.16\% & 84.63\% & 0.24  & 0.41  & 0.52  & 0.62 \\
    \textbf{0.4 } & \textbf{95.61\%} & \textbf{99.08\%} & \textbf{0.20} & \textbf{0.31} & \textbf{0.45} & \textbf{0.61} \\
    0.7   & 71.23\% & 78.16\% & 0.31  & 0.44  & 0.52  & 0.72 \\
    0.9   & 76.32\% & 79.59\% & 0.29  & 0.42  & 0.52  & 0.74 \\
    1.0   & 2.17\% & 2.82\% & 0.21  & 0.32  & 0.47  & 0.63 \\
    \hline
    \end{tabular}}%
  \label{tab:addlabel}%
\end{table}%

\subsection{Predicted Sequence}
In order to prove NGC-Seq2Seq provides better prediction than baselines, we compare the generated future poses of "discussion" provided by different models (i.e. Res-sup, Traj-GCN and NGC-Seq2Seq). According to Fig 5, Res-sup cannot predict the dynamics of body components. Prediction errors also exist with the Traj-GCN model after 800 ms. Although the NGC-Seq2Seq model also has slight differences compared with the ground truth after 800 ms, the differences are smaller than with the other methods, making it the most reliable approach. It is also worth noting that only joint-scaling was applied in training the models in this work making it a more challenging problem than when part-scaling is also considered.

\begin{figure}[thpb]
\centering
\noindent\includegraphics[width=\columnwidth]{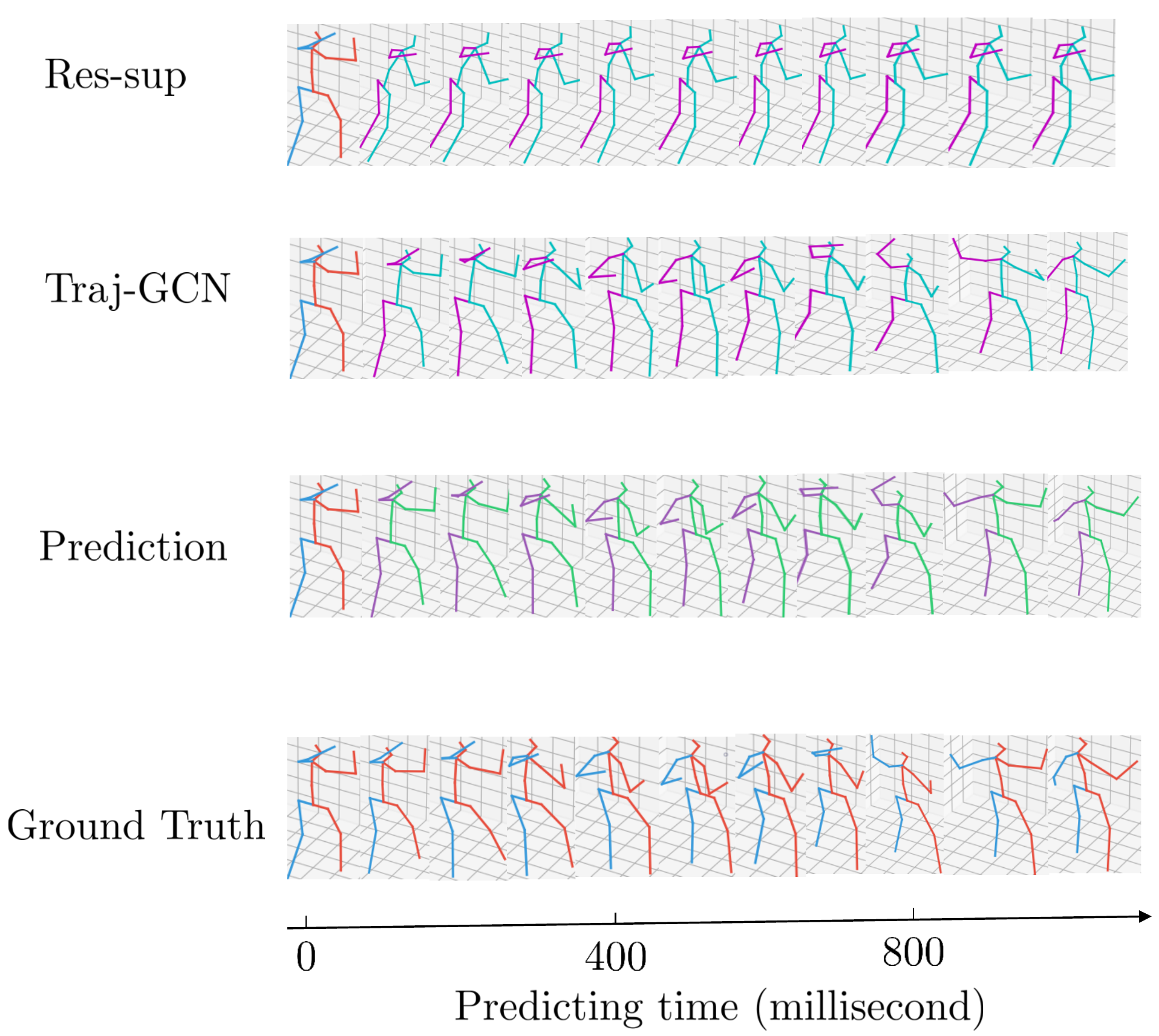}
\caption{Visualization of motion prediction on Human 3.6M: the predicted pose of "discussion" with NGC-Seq2Seq, Res-sup and Traj-GCN, compared to the ground truth.}
\label{fig11}
\end{figure}

\begin{table}[h]
  \centering
  \caption{Average cross-subject recognition accuracy and long-term MAE on NTU-RGB+D over  10 Monte Carlo simulations. $p$ values are also reported for t-tests comparing the ablated and full models.}
  \small\addtolength{\tabcolsep}{-8pt}
  \resizebox{\columnwidth}{!}{%
    \begin{tabular}{|c|c|c|c|c|}
    \hline
          & CS    & MAE(1000) & p-value(CS) & p-value(MAE) \\
    \hline
    Without Bi-LSTM & 87.96 & 1.52  & 1.26E-08 & 8.23E-06 \\
    \hline
    Without NGC & 72.59 & 1.77  & 3.56E-15 & 4.86E-07 \\
    \hline
    Our model & 95.61 & 1.41  & N/A & N/A \\
    \hline
    \end{tabular}}%
  \label{tab:addlabel}%
\end{table}%

\subsection{Statistical Significance}
In order to statistically validate the benefit of using Bi-LSTM and NGC attention on action recognition and long-term prediction, a Monte Carlo experiment was conducted in which models with and without Bi-LSTM and NGC were trained and tested on 10 random splits of the NTU-RGBD dataset. Table V illustrates the average accuracy and MAE of each model over the 10 splits. The $p$-values for a t-test for the differences between the mean accuracy and mean MAE of the ablated models and the full model are also reported. The t-test is a commonly used statistical test which evaluates whether or not the observed difference between the means of two sets of data is random or statistically significant.  The $p$-value for action recognition and long-term motion prediction are both close to zero ($p<<0.05$), confirming that the differences between models is statistically significant. Hence, we can conclude that the proposed model offers a significant increase in performance for action recognition and motion prediction over models without a Bi-LSTM or without NGC attention implemented.

\section{Conclusion}
In this paper, a non-local graph convolutional (NGC) network model is proposed to solve the challenges of capturing long-range dependencies and having a shared feature map for both human action recognition and motion prediction. The human motion prediction module consists of a NGC module and a LSTM decoder. The activity recognition module contains a couple of layers of NGC and scoring by conditional random fields. Results for both prediction and recognition show that the proposed model can capture inner spatial-temporal correlation in a task as well as long term dependencies, enabling it to achieve a consistent improvement over existing methods. Future work will look at incorporating the predicted human state as an input to a collaborative robot controller to enable it achieve enhanced safety and reduced execution times for HRC tasks.

\medskip
\small
\bibliographystyle{unsrt}
\bibliography{RGCN}
\end{document}